\documentclass[letterpaper, 10 pt, conference]{ieeeconf}  % Comment this line out if you need a4paper
\usepackage{booktabs}
\usepackage{multirow}
\usepackage{graphicx}
\usepackage{pifont}
\usepackage{hyperref}
\usepackage{amsmath}
\usepackage{amsfonts}
\usepackage{booktabs}
\usepackage{pifont}
\usepackage{todonotes}
\usepackage[dvipsnames]{xcolor}
\usepackage[table]{xcolor}
\usepackage[T1]{fontenc}
\usepackage{textcomp}
\definecolor{oursblue}{HTML}{DCE9F7}
\usepackage{tikz}

\DeclareRobustCommand{\stepicon}[1]{%
  \tikz[baseline=(char.base)]{
    \node[
      circle,
      fill=black,
      text=white,
      inner sep=1.3pt,
      font=\scriptsize\bfseries
    ] (char) {#1};
  }%
}

\definecolor{ourblue}{HTML}{016CFA}
\definecolor{ourred}{HTML}{FF0000}
\newcommand{\dpos}[1]{\textcolor{ourblue}{#1}}
\newcommand{\dneg}[1]{\textcolor{ourred}{#1}}
\newcommand{\dzero}[1]{\textcolor{gray}{#1}}

\IEEEoverridecommandlockouts                              % This command is only needed if 
\title{\LARGE \bf
INTERACT: \underline{Inter}active Planning for Autonomous Driving via \underline{A}nchor-\underline{C}onditioned Prediction and \underline{T}rust-Region Refinement
}

\author{Aron Distelzweig$^{1, 2}$, Andreas Look$^{3}$, Faris Janjo\v{s}$^{1}$, Steffen Hagedorn$^{1}$, Luigi Palmieri$^{1}$, Joschka Boedecker$^{2}$% <-this % stops a space
\thanks{$^1$Robert Bosch GmbH, Germany.}%
\thanks{$^2$University of Freiburg, Germany. }%
\thanks{$^3$Coburg University, Germany.}
}

\begin{document}

\maketitle
\thispagestyle{empty}
\pagestyle{empty}

%%%%%%%%%%%%%%%%%%%%%%%%%%%%%%%%%%%%%%%%%%%%%%%%%%%%%%%%%%%%%%%%%%%%%%%%%%%%%%%%
\begin{abstract}
Driving in dense urban traffic is interactive: whether a merge or an unprotected
turn succeeds depends on how surrounding agents respond to the ego vehicle.
Conventional planners predict first and plan second and, therefore, cannot account
for this dependency. Methods that integrate prediction and planning either train
both jointly, which introduces task interference, or keep them separate and are
restricted to a predefined set of proposals. We present \textsc{Interact}:
\underline{Inter}active Planning for Autonomous Driving via
\underline{A}nchor-\underline{C}onditioned Prediction and
\underline{T}rust-Region Refinement. Our key insight is that surrounding agents
react to the intent a trajectory expresses rather than to its exact realization,
so a single reactive prediction stays valid across an entire family of plans.
\textsc{Interact} therefore decomposes interactive planning into prediction
across driving intents and optimization within each intent. We derive a small set
of diverse intents, which we call anchors, from map geometry, query a dedicated
ego-conditioned prediction model once per anchor, and refine every anchor with
the Cross-Entropy Method under a trust-region penalty that keeps the refined plan
close enough to its anchor for the conditioned reaction to still apply.
Prediction thus remains a separate model, avoiding task interference, while
conditioning on anchors preserves the dependency. Because each anchor is refined
continuously, the final plan is not restricted to the anchor set, yet
\textsc{Interact} requires only one predictor query per anchor rather than one
per candidate plan, with all anchors processed in parallel. On the nuPlan and
interPlan closed-loop benchmarks, \textsc{Interact} sets a new state of the art,
with the largest gains precisely in the interactive scenarios that motivate the
method. The code will be released upon acceptance.
\end{abstract}
\section{Introduction}
Driving in dense urban traffic is inherently interactive: at unprotected left
turns, lane merges, or passages through narrow gaps, the feasibility of a
maneuver depends on how surrounding agents respond to the ego vehicle. A
planner whose forecast of the other agents is independent of its own plan
implicitly assumes that no one will react, which yields overly conservative
behavior and, in the limit, the frozen-robot problem~\cite{trautman2010unfreezing,
hagedorn2024review}. No amount of forecasting accuracy resolves this; the
deficit lies in the missing coupling. Integrated
prediction and planning, in which the prediction is conditioned on the ego plan,
is the natural remedy.

However, on the nuPlan benchmark~\cite{caesar2022nuplan}, the strongest planners that
reason about surrounding agents still do so through constant-velocity
forecasts~\cite{Dauner2023PDM}, whereas approaches built on reactive
predictions have not converted their more expressive interaction modeling into
better closed-loop driving~\cite{huang2024dtpp,huang2023gameformer, distelzweig2025perfectprediction}. Two properties of current approaches account for this.

First, learned integration approaches commonly train prediction and planning jointly within a single
network~\cite{huang2024dtpp,huang2023gameformer}, but the two are distinct tasks
that interfere under a shared parameterization~\cite{seo2026predictionplanning},
leaving the predictions behind dedicated state-of-the-art models and degrading
planning performance as well. The plan is additionally regressed from
expert demonstrations~\cite{huang2023gameformer, zhong2025coplanner,
yu2025hype}, which inherits the statistics of driving data where
interactive situations are rare and lane keeping dominates, so the imitation
objective is dominated by the very regime in which interaction reasoning is
irrelevant. Evaluated in interactive scenarios, the resulting proposal sets fall
below a simple lane-following baseline~\cite{distelzweig2025perfectprediction}.

\begin{figure}
    \includegraphics[width=\columnwidth]{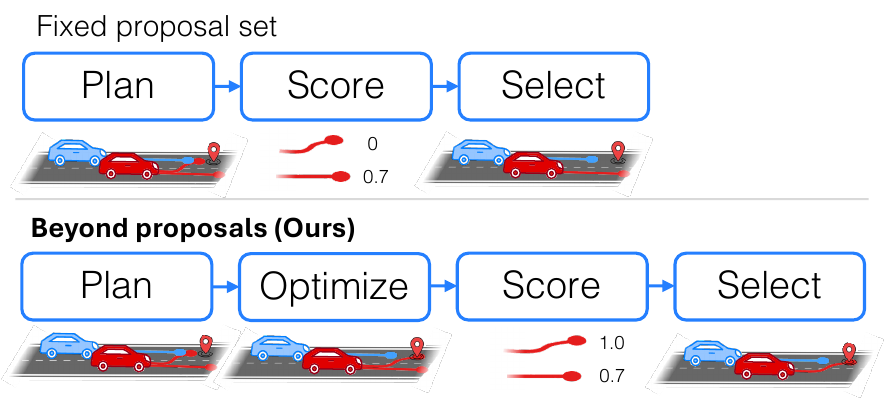}
    \vspace{-20pt}
\caption{\textbf{Interactive planning beyond a fixed proposal set.}
Proposal-based planners (top) score a fixed set of candidate trajectories and
select the best one; the executed plan can therefore never be better than the
best candidate in that set. \textsc{Interact} (bottom) treats each candidate as
an \emph{anchor} that expresses a driving intent, queries an ego-conditioned
predictor once per anchor, and refines the anchor continuously under a
trust-region penalty that keeps the reaction valid. Scoring and selection then
operate on refined plans, so the final trajectory is no longer confined to the
initial set, at the cost of one predictor query per anchor rather than one per
candidate plan.}
    \label{fig:fig1}
\vspace{-15pt}
\end{figure}

Second, approaches that avoid learning the plan instead rely on a fixed set of
proposals, scored against their conditioned
predictions~\cite{distelzweig2025biber, huang2024dtpp}, and are thereby confined
to whatever that set contains. The proposals are generated before any prediction
is available, so conditioning determines which plan is selected but never what
any plan looks like. Enlarging the set is no remedy, as a separate predictor
must be queried once for every plan it is conditioned on.

We therefore approach the integration of prediction and planning differently.
Our key insight is that surrounding agents primarily react to the intent  rather than to the exact realization. Trajectories that share an intent induce near-identical
reactions, so a prediction conditioned on one representative trajectory stays
valid while that trajectory is refined locally. A single query therefore covers
an entire family of plans, which is what makes a dedicated
predictor affordable in closed loop. This lets us decompose interactive planning
into prediction across intents and optimization within each intent, and to
design the two stages independently with methods suited to their respective
objectives.

We first derive a small set of diverse intents, which we call anchors, from map
geometry. Unlike rule-based proposal sets, anchors are not the final output but
the conditioning variable of the predictor: each one is subsequently optimized
against its own forecast. We then adapt a state-of-the-art multi-agent prediction model into an
anchor-conditioned predictor, queried once per planning step to return one
reactive scene prediction per anchor. Because this model is trained for
prediction alone, it retains the quality of a dedicated predictor. Finally, we
refine each anchor with the Cross-Entropy Method (CEM)~\cite{Rubinstein1999CEM}
against its conditioned prediction, adding a trust-region term that keeps the
refined plan close enough to its anchor for the conditioned reaction to still
apply. The predictor is therefore never re-queried during optimization. The same
construction removes a known weakness of CEM in multimodal settings, where a
single search distribution tends to average across distinct modes: here the
anchors supply the modes, and the trust region confines each refinement to the
one it started from. 

\noindent In summary, our contributions are:
\begin{itemize}
  \item We formulate interactive planning as prediction across intents and
        optimization within intents, based on the insight that the reaction of
        surrounding agents is determined by the intent a trajectory expresses
        rather than by its exact realization.
  \item We turn a state-of-the-art multi-agent prediction model into an
        anchor-conditioned predictor that produces one reactive scene prediction
        per anchor, keeping prediction and planning as separate
        specialized components.
  \item We introduce a trust-region CEM refinement that optimizes a novel continuous
relaxation of the PDM score, keeps each plan within the range in which its
conditioned prediction remains valid, mitigates mode averaging, and runs
independently across anchors.
  \item \textsc{Interact} sets a new state of the art on nuPlan and interPlan,
        with the largest gains in interactive scenarios.
\end{itemize}
\section{Related Work}

\subsection{Integrated Prediction and Planning}
To date, the predominant order of tasks in modular Autonomous Driving pipelines is perception followed by prediction, planning, and control.
Even when optimized end-to-end for the final driving objective, this design remains inherently reactive: the ego vehicle can react to the predicted behavior of other traffic participants but the predictions do not account for the planned ego behavior.
Instead of realistically modeling interactions, the ego vehicle disregards its own influence on the traffic scene and behaves overly conservative~\cite{Dauner2023PDM, hagedorn2024review}.
Planning the ego trajectory first and then conditioning prediction on it does not resolve the issue as it barely inverts the dependency while keeping it unidirectional and therefore reactive~\cite{song2020pip}.
Joint prediction and planning goes beyond reactive modeling by solving both tasks in a common step.
However, it assumes that all vehicles follow a joint driving objective, resulting in a single global solution~\cite{huang2023gameformer, chekroun2023mbappe}.
Only bidirectional integration schemes, where prediction is conditioned on planning and vice versa explicitly model interactions by rolling out a what-if simulation of alternative futures to ultimately decide for an ego behavior.
While conceptually promising, bidirectional interaction modeling comes at a high computational cost.
Existing methods take different measures to limit this cost:
TPP~\cite{chen2023tpp} and DTPP~\cite{huang2024dtpp} perform a tree search with strictly restricted depth whereas HYPE~\cite{yu2025hype} constrains its width. HPP~\cite{liu2025hybrid} alternates prediction and ego plan optimization but starts from a joint prediction instead of considering multimodal futures. In contrast, CoPlanner~\cite{zhong2025coplanner} makes a conservative unimodal short-term prediction and only accounts for diverse behaviors in the long term whereas BIBeR~\cite{distelzweig2025biber} asserts a game-theoretic iterative best response framework.
Our work instead avoids the cost of full bidirectional rollouts by a simplified formulation: interactive planning as prediction across discrete intents and optimization within those intents.
This design is grounded in the assumption that surrounding agents primarily react to the overarching intent, allowing us to reduce complexity in a task-compliant manner.

\subsection{Conditional Prediction}
General trajectory prediction models have achieved high forecasting accuracy~\cite{liu2024laformer, wu2024smart, zhou2022hivt}. However, the benefit of these models for interactive planning remains marginal~\cite{distelzweig2025perfectprediction}.
One limitation is that most existing models are unconditional, even though this conditioning is crucial for modeling bidirectional interactions.
They forecast future behavior based only on historical context and do not account for the intended future behavior of the ego vehicle.
While integrated approaches like DTPP~\cite{huang2024dtpp} and HYPE~\cite{yu2025hype} address this by learning custom conditional prediction models, these architectures share model capacity with the planning objective and often fall short of the accuracy achieved by dedicated prediction models.
To overcome this, we adopt SMART~\cite{wu2024smart} as our base model and adapt it into an ego-conditioned predictor.
As detailed in the following section, this integration leverages the model's autoregressive structure, requiring no architectural modifications or retraining, which ensures that baseline prediction performance is preserved.

\subsection{Test Time Optimization}
Test-time trajectory optimization refines behavioral proposals prior to execution.
Diffusion-ES~\cite{yang2024diffusiones} employs a similar optimization concept to ours but faces computational inefficiencies due to the iterative querying of a diffusion model.
Furthermore, it assumes constant-velocity predictions, thereby failing to model interactions between the ego vehicle and surrounding agents.
Another recent method~\cite{xu2026toad} optimizes trajectories for end-to-end architectures but does not integrate reactive multi-agent predictions into the refinement process.
In contrast, our approach utilizes the conditional predictions outlined in the previous section to explicitly model bidirectional interactions.
We apply Cross-Entropy Maximization (CEM) to independently refine distinct maneuver anchors against their respective reactive scene predictions.
To circumvent the computational burden of repeatedly querying the SMART predictor during this search, we constrain the refinement to a limited trust region.
This keeps the optimization local, ensuring the refined trajectory remains within a valid neighborhood of the initial prediction anchor and naturally mitigating the mode-averaging problem associated with CEM.
Unlike existing approaches that rely on discrete, rule-based cost functions~\cite{Dauner2023PDM, hagedorn2026diffusearch}, we evaluate candidate trajectories using a continuous surrogate score.
This continuous formulation resolves the flat plateaus inherent to discrete scoring metrics, enabling smoother convergence and a more precise ranking of interactive behaviors.

\section{Methodology}

\begin{figure*}[htbp]
    \centering
    \includegraphics[width=\linewidth]{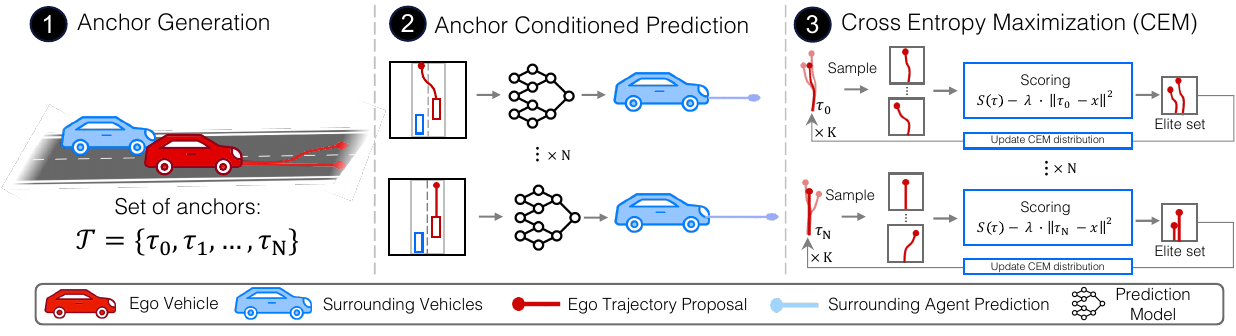}
    \caption{Overview of INTERACT. \stepicon{1} \textbf{Anchor Generation:} We generate a diverse set of anchors, representing different ego intents. \stepicon{2} \textbf{Anchor-Conditioned Prediction:} For each anchor, a conditional prediction model predicts how surrounding agents react to the corresponding ego intent. \stepicon{3} \textbf{Trajectory Refinement:} Each anchor is independently refined with Cross-Entropy Maximization (CEM) within a local region, preserving its maneuver intent and thereby the validity of the corresponding anchor-conditioned prediction.}
    \label{fig:approach}
    \vspace{-10pt}
\end{figure*}

Our approach consists of four steps. First, we generate a set of diverse, kinematically feasible ego trajectories from route geometry, called anchors (Sec.~\ref{sec:anchors}). Second, for each anchor, a prediction model predicts how the surrounding agents react to this ego intent (Sec.~\ref{sec:prediction}). Third, we refine each anchor against its own prediction using CEM~\cite{Rubinstein1999CEM} (Sec.~\ref{sec:cem}). Fourth, a pairwise comparison selects the plan that is executed (Sec.~\ref{sec:selection}). Fig.~\ref{fig:approach} provides an overview.

\subsection{Anchor Generation}
\label{sec:anchors}
Anchors are ego trajectories that represent a driving maneuver. They are used to condition the prediction and to initialize the refinement. The anchor set is constructed from map geometry: each path is combined with several target speeds. The paths are the route centerline and the neighboring lanes. Each of them is not followed directly, since the ego is generally offset from it in both position and heading. For a point $\ell_m = (x_m, y_m, \psi_m)$ on the lane, a cubic B\'ezier curve connects the current pose $(x_0, y_0, \psi_0)$ to it, yielding a path $\Gamma$ parameterized by arc length $s$. Here, $(x, y)$ denotes the position and $\psi$ the heading. Varying the point yields the same maneuver at different levels of urgency: an early merge produces a fast and a late one a slow lane change. Every path is followed with five different speed profiles. Given the current speed $v_0$ and the local speed limit $v_{\lim}$, profile $k$ targets $v^{(k)} = \kappa_k v_{\lim}$ with a fixed fraction $\kappa_k$ and approaches it at a constant rate, 
\begin{equation} v_{t+1} = \begin{cases} \min(v_t + a_{\mathrm{acc}} \Delta t,\; v^{(k)}), & v_t < v^{(k)},\\[2pt] \max(v_t - a_{\mathrm{dec}} \Delta t,\; v^{(k)}), & \text{otherwise}, 
\end{cases} 
\label{eq:speed_profile} 
\end{equation} 
\begin{equation} s_{t+1} = s_t + v_{t+1} \Delta t , \label{eq:arclength} 
\end{equation} starting at the point on $\Gamma$ closest to the ego vehicle. The reference poses then follow from the path geometry as $\bar{p}_t = \Gamma(s_t)$. Path and speed profile thus separate the two degrees of freedom: the path determines the shape of the maneuver, while the profile determines its temporal progression. The resulting reference trajectories are tracked by an LQR controller, yielding control sequences that are rolled out with the kinematic bicycle model $\mathrm{BM}(\cdot)$. This produces dynamically feasible trajectory candidates. The candidates are ranked by the score of Sec.~\ref{sec:cem}. Finally, $N$ anchors \begin{equation} \mathcal{T} = \{\tau_1, \dots, \tau_N\}, \qquad \tau_i = (p_1^i,\dots,p_T^i), \label{eq:anchorset} \end{equation} are selected by farthest-point sampling in trajectory space, initialized with the highest-scoring candidate rather than a random one. $p_t^i = (x_t^i,y_t^i,\psi_t^i)$ denotes the pose of anchor $i$ at time step $t$. We additionally retain the corresponding control sequence $u_i^{\mathrm{anc}}$ that generated each selected anchor. Sampling for spatial diversity rather than for score ensures that the subsequent optimization starts from genuinely different driving intents instead of $N$ variants of the same one.

% --------------------------------------------------------------------------------------------------------------

\subsection{Anchor-Conditioned Prediction}
\label{sec:prediction}
We use SMART~\cite{wu2024smart}, which is a next-token prediction model that represents each agent trajectory as a sequence of discrete motion tokens and factorizes the joint future autoregressively over time steps. Each token encodes a short trajectory segment of fixed duration. The model is trained with teacher forcing on ground-truth token sequences. Writing $a_t^{\,n}$ for the token of agent $n$ at time step $t$, $N_a$ for the number of agents and $\mathcal{C}$ for the scene context (map and observed history), the model defines
\begin{equation}
  p_\theta\bigl(a_{1:T}^{\,1:N_a} \mid \mathcal{C}\bigr)
  =
  \prod_{t=1}^{T} \prod_{n=1}^{N_a}
  p_\theta\bigl(a_t^{\,n} \mid a_{<t}^{\,1:N_a}, \mathcal{C}\bigr),
  \label{eq:autoregressive}
\end{equation}
i.e. the factorization is causal in time while agents within one step are conditionally independent given the joint history. Rollouts are obtained by sampling: at each time step every agent draws a token from its predicted distribution, all tokens are written back into the sequence, and the interaction structure is rebuilt from the updated poses before next step.

We exploit this structure to condition the prediction on anchors. Let $n = 0$ denote the ego. For anchor $\tau_i$ we tokenize its pose sequence into $\bar{a}_{1:T}^{\,0}(\tau_i)$ using the same tokenizer, and replace the ego tokens during sampling in Eq.~\ref{eq:autoregressive}:
\begin{equation}
  p_\theta\bigl(a_t^{\,n} \mid a_{<t}^{\,1:N_a}, \mathcal{C}, \tau_i\bigr)
  =
  \begin{cases}
    \mathbb{I}\bigl[a_t^{\,0} = \bar{a}_t^{\,0}(\tau_i)\bigr], & n = 0,\\[3pt]
    p_\theta\bigl(a_t^{\,n} \mid a_{<t}^{\,1:N_a}, \mathcal{C}\bigr), & n \neq 0 .
  \end{cases}
  \label{eq:conditioning}
\end{equation}
Concretely, after each decoding step the ego's sampled token is overwritten by the anchor token, so that all subsequent predictions are conditioned on the anchor. Since the model factorizes autoregressively over the joint sequence of all agents, this substitution amounts to exact conditioning on $\tau_i$ rather than to sampling the ego freely. Reading out only the non-ego tokens then gives the conditioned prediction
\begin{equation}
  \hat{Y}_i = \bigl(\hat{y}_{t}^{\,n}\bigr)_{\,n = 1 \dots N,\; t = 1 \dots T}
  \sim p_\theta\bigl(\cdot \mid \mathcal{C}, \tau_i\bigr),
  \label{eq:prediction}
\end{equation}
where $\hat{y}_t^{\,n}$ denotes the predicted state of agent $n$ at time step $t$, $n{=}0$ indexes the ego vehicle and $n{=}1, \dots, N$ the surrounding agents, $\mathcal{C}$ is the scene context and $\tau_i$ the $i$-th ego anchor. All $N$ conditioned rollouts are computed in parallel.

% --------------------------------------------------------------------------------------------------------------
\subsection{Refinement}
\label{sec:cem}
\begin{figure}[t]
\centering
\includegraphics[width=\columnwidth]{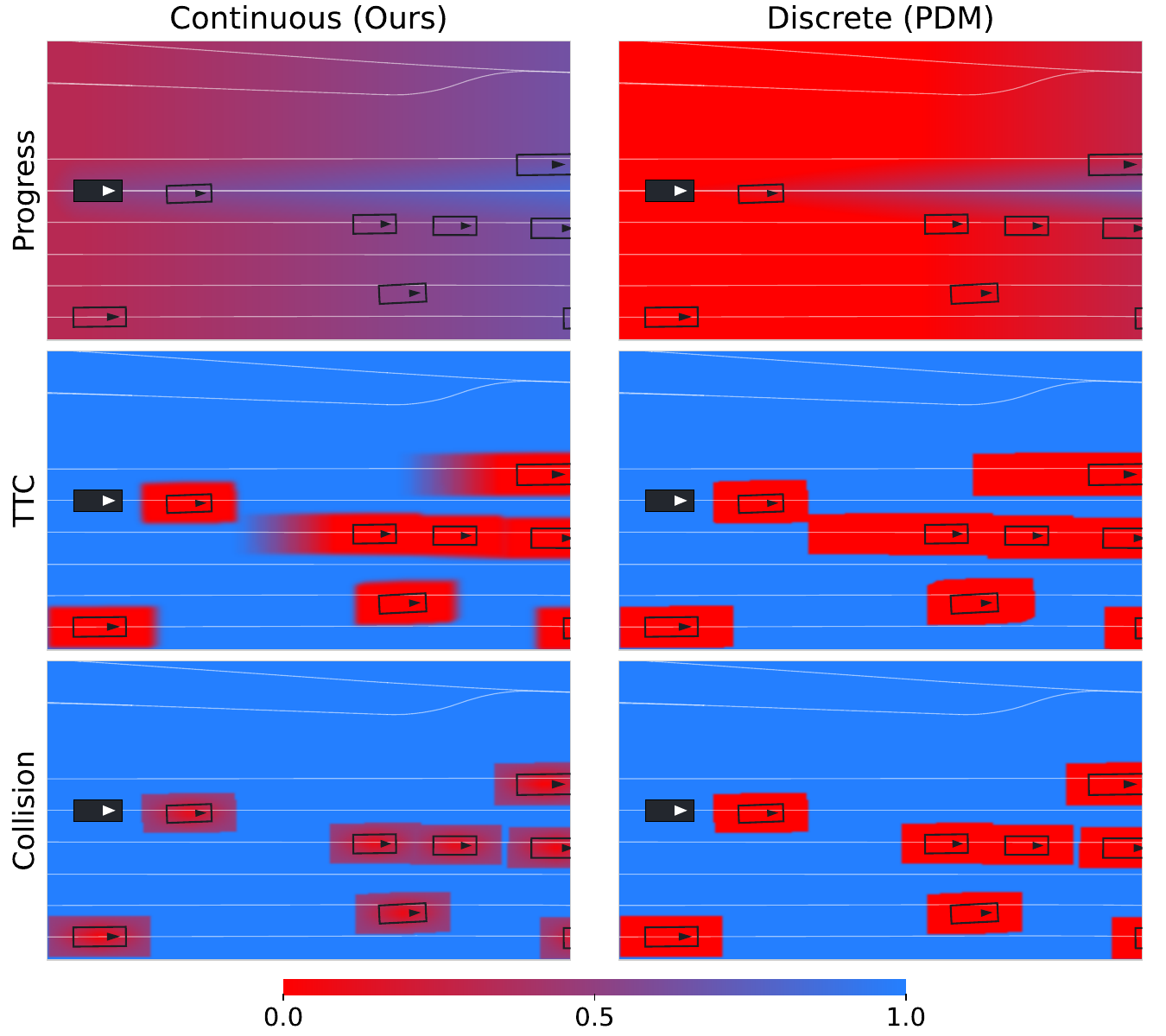}
\vspace{-18pt}
\caption{Comparison of our continuous score components (left) with their discrete PDM counterparts (right) on a nuPlan scenario. For every map location, the progress, TTC, and collision terms are evaluated. Color encodes the score value. The discrete terms are piecewise constant: gradients exist only at edges, leaving the search without guidance on the flat plateaus, backing up scores identically to standing still, and every colliding pose is equally bad. The continuous terms provide an informative gradient throughout: progress degrades smoothly behind the ego and pulls laterally toward the goal lane, TTC decays with the closing distance to slower vehicles, and the collision term grades severity by footprint overlap, pointing out of the conflict.}
\label{fig:score}
\vspace{-10pt}
\end{figure}
Each anchor $\tau_i$ is refined independently against its conditioned
prediction $\hat{Y}_i$, which is fixed during refinement. We optimize in the space of control sequences
$u=(u_t)_{t=1}^{T}$ with $u_t=(a_t,\dot{\delta}_t)$, where $a_t$
denotes the acceleration and $\dot{\delta}_t$ the steering rate.
A control sequence is rolled out with the kinematic bicycle model,
\begin{equation}
    \mathrm{BM}(u) = (p_1,\dots,p_T),
    \label{eq:rollout}
\end{equation}
where $p_t=(x_t,y_t,\psi_t)$ denotes the resulting ego pose at step $t$.
For each $\tau_i$, we define the objective to be maximized
\begin{equation}
    J_i(u)
    =
    S\bigl(\mathrm{BM}(u),\hat{Y}_i\bigr)
    -
    \lambda \frac{1}{T}
    \sum_{t=1}^{T}
    \left\|
        (x_t,y_t)-(x_t^i,y_t^i)
    \right\|_2 .
    \label{eq:objective}
\end{equation}

Here, $(x_t,y_t)$ denote the positions obtained by rolling out $u$ with $\mathrm{BM}$, while $(x_t^i,y_t^i)$ are the corresponding positions of anchor $\tau_i$.
The first term evaluates the candidate according to general planning criteria, including comfort, collision avoidance, and drivable-area compliance, where interaction-dependent criteria are evaluated against the predicted agent trajectories $\hat{Y}_i$. The second term penalizes deviations from the anchor. The score $S(\cdot)$ is a continuous surrogate of the PDM score~\cite{Dauner2023PDM}, preserving its original multiplicative-gate and weighted-average structure. We replace the discrete collision, drivable-area, driving-direction, and time-to-collision (TTC) terms with continuous counterparts and use continuous progress instead of its gated variant. We show a comparison in Fig.~\ref{fig:score}. This yields a smoother ranking of candidate trajectories and simplifies the optimization. We optimize Eq.~\ref{eq:objective} using the Cross-Entropy Method (CEM)~\cite{Rubinstein1999CEM}, maintaining per anchor a diagonal
Gaussian distribution with parameters $\mu_i^{(k)}, \sigma_i^{(k)} \in \mathbb{R}^{T\times 2}$ over control sequences. Since the control sequence of each anchor is retained during anchor generation, the search is initialized directly at the anchor,
\begin{equation}
\mu_i^{(0)} = u_i^{\mathrm{anc}}, \qquad
\sigma_{i,t}^{(0)} = \rho \cdot \tfrac{1}{2}\bigl(u_{\max}-u_{\min}\bigr),
\quad t=1,\dots,T,
\label{eq:warmstart}
\end{equation}
where $u_i^{\mathrm{anc}}$ are the anchor controls, $u_{\min},u_{\max}\in\mathbb{R}^2$ are fixed control bounds, and $\rho=0.15$ sets the initial exploration range, scaling by the admissible range accounts for the different magnitudes of acceleration and steering rate.

In each iteration $k$ we draw a population of $P$ control sequences,
\begin{equation}
u_i^{(k,j)} = \operatorname{clip}\!\bigl(\mu_i^{(k)}
  + \sigma_i^{(k)} \odot \varepsilon^{(k,j)}\bigr),
\qquad j=1,\dots,P,
\label{eq:sampling}
\end{equation}
where $\odot$ is the elementwise product and clipping keeps the samples inside the control bounds. Following iCEM~\cite{PinneriiCEM}, the perturbations $\varepsilon^{(k,j)}\in\mathbb{R}^{T\times 2}$ are unit-variance colored noise with power spectral density $\propto 1/f^{\beta}$ ($\beta{=}2$), drawn per control dimension over the full horizon: each step is marginally Gaussian, but consecutive steps are correlated, which concentrates the search on smooth control sequences rather than white-noise chatter.

Every candidate is scored by Eq.~\ref{eq:objective} and the $E$ best form the elite set $\mathcal{E}_i^{(k)}$, to which the distribution is refit,
\begin{equation}
\mu_i^{(k+1)} = \operatorname{mean}\bigl(\mathcal{E}_i^{(k)}\bigr),
\qquad
\sigma_i^{(k+1)} = \operatorname{std}\bigl(\mathcal{E}_i^{(k)}\bigr).
\label{eq:refit}
\end{equation}
After $K$ iterations, the best control sequence $u_i^*$ found during the search defines the refined trajectory
\begin{equation}
\tau_i^* = \mathrm{BM}(u_i^*).
\label{eq:refined_trajectory}
\end{equation}

% --------------------------------------------------------------------------------------------------------------
\subsection{Final Plan Selection}
\label{sec:selection}
The final plan is given by
\begin{equation}
  \tau^{\star} = \operatorname*{arg\,max}_{i = 1 \dots N}\;
  S\bigl(\tau_i^{*}, \hat{Y}_i\bigr).
  \label{eq:argmax}
\end{equation}

The trust region confines each refinement to the validity region of its own prediction, so no re-prediction is required during the $K$ CEM iterations.
Beyond this, the per-anchor process of warm start, sampling, rollout and scoring is fully independent across anchors and therefore parallelizable, with only the final plan selection requiring synchronization.
\section{Experiments}
\begin{table*}[tp]
    \centering
    \footnotesize
    \setlength{\tabcolsep}{3pt}
    \resizebox{\textwidth}{!}{%
    \begin{tabular}{llrrrrrrr}
        \toprule
        Predictions & Method
        & Val14 (R) $\uparrow$
        & Test14-hard (R) $\uparrow$
        & Test14-random (R) $\uparrow$
        & Dynamic-10 (R) $\uparrow$
        & interPlan (R) $\uparrow$
        & interPlan-LC (R) $\uparrow$
        & interPlan-full (R) $\uparrow$ \\
        \midrule

        % NO PREDICTIONS
        \multirow{7}{*}{None}
        & Urban Driver~\cite{scheel2022urban}
        & 64.11 & 49.95 & 67.15 & 65.98 & 4.00 & 9.65 & --- \\
        & GC-PGP~\cite{hallgarten2023prediction}
        & 63.82 & 39.36 & 51.39 & --- & 14.55 & 34.99 & --- \\
        & PlanCNN~\cite{renz2022plant}
        & 72.00 & 52.20 & 67.50 & --- & --- & 58.35 & --- \\
        & PlanTF~\cite{jcheng2023plantf}
        & 76.95 & 61.61 & 79.58 & --- & 30.53 & 55.18 & --- \\
        & PDM-Open~\cite{Dauner2023PDM}
        & 54.24 & 35.83 & 57.23 & --- & --- & 23.35 & --- \\
        & DiffusionPlanner~\cite{zheng2025diffusion}
        & 82.80 & 69.22 & 82.93 & --- & 25.76 & 25.96 & --- \\
        & TerraZero~\cite{wu2026terrazero}
        & \textbf{94.19} & --- & --- & --- & \textbf{70.87} & --- & 71.31 \\

        \midrule

        % CONSTANT VELOCITY
        \multirow{3}{*}{Constant Velocity}
        & PDM-Closed~\cite{Dauner2023PDM}
        & 92.12 & 75.19 & 91.63 & \underline{91.50} & 41.88 & 61.55 & --- \\
        & PDM-Hybrid~\cite{Dauner2023PDM}
        & 92.11 & 76.07 & 91.28 & --- & 41.61 & 61.72 & --- \\
        & Diffusion-ES~\cite{yang2024diffusiones}
        & 92.00 & --- & --- & --- & --- & --- & --- \\

        \midrule

        % REACTIVE LEARNED
        \multirow{7}{*}{Reactive Learned}
        & GameFormer~\cite{huang2023gameformer}
        & 79.78 & 67.05 & 82.05 & --- & 21.26 & 30.49 & --- \\
        & DTPP~\cite{huang2024dtpp}
        & 66.09 & 63.66 & --- & 89.78 & 30.32 & 67.88 & --- \\
        & HYPE~\cite{yu2025hype}
        & 92.63 & --- & --- & 91.27 & --- & --- & --- \\
        & CoPlanner~\cite{zhong2025coplanner}
        & 93.13 & 78.59 & 92.00 & --- & --- & --- & --- \\
        & BIBeR-CV*~\cite{distelzweig2025biber}
        & 92.75 & \underline{83.00} & \textbf{93.01} & --- & \underline{60.05} & \underline{84.53} & --- \\
        & BIBeR~\cite{distelzweig2025biber}
        & 89.79 & 79.57 & 90.92 & --- & 55.59 & 81.17 & --- \\

        & \cellcolor{oursblue}\textbf{INTERACT (Ours)}
        & \cellcolor{oursblue} \underline{93.18}
        & \cellcolor{oursblue}\textbf{85.28}
        & \cellcolor{oursblue}\underline{92.48}
        & \cellcolor{oursblue}\textbf{94.65}
        & \cellcolor{oursblue} 58.29
        & \cellcolor{oursblue}\textbf{89.07}
        & \cellcolor{oursblue} \textbf{79.11} \\

        \bottomrule
    \end{tabular}%
    }
    \caption{Comparison of planning performance on the nuPlan benchmarks, grouped by the type of agent prediction the planner consumes. We report closed-loop reactive (R) scores for the Val14, Test14-hard, Test14-random, and Interactive splits, as well as for interPlan, interPlan-LC, and interPlan-full. Higher values indicate better performance. Best and second-best values per column are marked in \textbf{bold} and \underline{underlined}, respectively. Our method is \colorbox{oursblue}{highlighted}. *BIBeR-CV filters candidate plans based on constant-velocity predictions but uses reactive learned predictions afterwards.}
    \vspace{-0.2cm}
    \label{tab:performance}
\vspace{-10pt}
\end{table*}

% ------------------------------------------------------------------------------------------------------------------------------
\subsection{Experimental Setup}
We build on nuPlan~\cite{caesar2022nuplan}, a closed-loop simulation framework based on real-world driving logs, and evaluate our approach on five established and widely used benchmarks. Val14~\cite{Dauner2023PDM} contains 1{,}118 scenarios from the nuPlan validation split, with up to 100 instances for each of 14 scenario types. Test14~\cite{jcheng2023plantf} comprises two variants: Test14-random uniformly samples scenarios from each type, whereas Test14-hard is constructed by evaluating 100 scenarios per type with the state-of-the-art rule-based planner PDM-Closed~\cite{Dauner2023PDM} and retaining the 20 lowest-scoring scenarios per type. Following~\cite{huang2024dtpp,yu2025hype}, we additionally evaluate on a split containing only dynamic scenario types that involve interactions with surrounding traffic. The split comprises 10 scenario types with 20 scenarios each, resulting in 200 scenarios in total. We refer to this split as Dynamic-10. Finally, interPlan~\cite{hallgarten2024interplan} consists of hand-crafted out-of-distribution scenarios covering challenging interactions such as maneuvering around parked vehicles, reacting to jaywalkers, and changing lanes in dense traffic. We report results on the commonly used subset of 80 scenarios, denoted as interPlan, and on the complete set of 335 scenarios, denoted as interPlan-full. In addition, we evaluate on a subset of 30 lane-change scenarios, denoted as interPlan-LC, which emphasizes ego-traffic interaction and is evenly split across low, medium, and high traffic densities.

% ------------------------------------------------------------------------------------------------------------------------------
\subsection{Implementation Details}
We extract $N=5$ anchors which span a time horizon of $4\,\mathrm{s}$. For the conditioned prediction, we transform the anchors into a sequence of $8$ tokens with each spanning a horizon of $0.5\,\mathrm{s}$.
CEM operates on sequences of acceleration and steering-rate commands over a $4\,\mathrm{s}$ horizon at $10\,\mathrm{Hz}$ ($T=40$). For each of the $N=5$ anchors we run $K=10$ iterations with a population of $P=24$ candidates and $E=6$ elites. The trust-region weight is $\lambda=0.1$ and $\lambda=0$ for anchors which lead to a collision under the conditioned prediction.

% ------------------------------------------------------------------------------------------------------------------------------
\subsection{Main Results}
\label{sec:main-results}

Tab.~\ref{tab:performance} compares our method (INTERACT) against existing planners, grouped according to the type of agent prediction used during planning. We distinguish three categories. \emph{None} do not explicitly model the future motion of surrounding agents and must therefore account for interactions implicitly. \emph{Constant-velocity} methods use fixed predictions that extrapolate the current motion of other agents independently of the candidate ego plan. Such predictions provide a simple  approximation and can be effective when interactions are limited. However, they cannot represent how other agents may react to different ego behaviors. Finally, \emph{reactive learned} methods explicitly model future agent behavior while accounting for the interaction with the ego vehicle. This is important in interactive scenarios, where different ego plans can lead to different responses from surrounding agents.

INTERACT achieves the best performance on Test14-hard, Dynamic-10, interPlan-LC, and interPlan-full, with closed-loop reactive (R) scores of $85.28$, $94.65$, $89.07$, and $79.11$, respectively, and ranks second on Val14 ($93.18$) and Test14-random ($92.48$). The gains are particularly pronounced on the interaction-focused benchmarks. On the Dynamic-10 split, INTERACT improves over the strongest reported method, PDM-Closed, by $+3.15$ and over the strongest competing reactive learned method, HYPE, by $+3.38$. On Test14-hard, INTERACT improves over the strongest reactive learned baseline, BIBeR-CV, by $+2.28$. Similarly, on interPlan-LC, which contains challenging lane-change interactions, INTERACT improves over BIBeR-CV by $+4.54$. The consistent gains across these interaction-heavy benchmarks indicate that explicitly accounting for how surrounding agents may react to candidate ego plans becomes increasingly beneficial as the degree of interaction increases.

A different baseline is TerraZero, which reaches the highest score on Val14 ($94.19$, $+1.01$ over INTERACT) and on interPlan ($70.87$), but is outperformed by INTERACT on interPlan-full by $+7.80$. Rather than improving the prediction consumed by a planner, TerraZero scales policy learning itself through self-play, which requires substantially more training interaction and yields a policy not decomposable into an explicit prediction and planning stage. The two directions are therefore orthogonal, and the strong interPlan-full result of INTERACT suggests that explicit reactive predictions remain competitive with and complementary to large-scale learned policies. We note that TerraZero results are not reported on all splits, which limits a full direct comparison.

On the less interaction-focused Test14-random split, INTERACT remains highly competitive, although the gains over approaches with fixed predictions are smaller than on the interaction-rich benchmarks. This highlights an important trade-off between fixed and reactive predictions. When the future motion of surrounding agents can be approximated sufficiently well without accounting for their response to the ego vehicle, simple fixed predictions can provide a strong and robust planning signal without introducing errors from a learned interaction model. In contrast, their underlying independence assumption becomes restrictive in strongly interactive situations. The consistent improvements of INTERACT on the Dynamic-10, Test14-hard, interPlan-LC, and interPlan-full splits therefore suggest that its main advantage lies precisely in these scenarios, where reasoning about the coupled future behavior of the ego vehicle and surrounding traffic is most important.

% ------------------------------------------------------------------------------------------------------------------------------
\subsection{Ablation Studies}

\begin{figure}[t]
\centering
\includegraphics[width=.9\columnwidth]{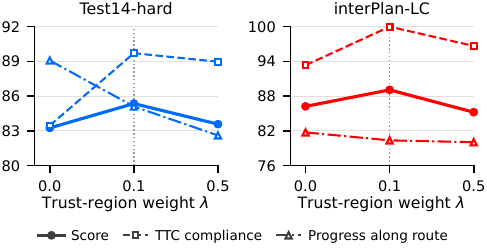}
\vspace{-10pt}
\caption{Sensitivity to the trust-region weight $\lambda$ on Test14-hard (left) and interPlan-LC (right): closed-loop score (solid), TTC compliance (dashed) and progress along the route (dash-dotted). A small $\lambda$ lets the CEM trade safety for progress, a large $\lambda$ keeps it close to the anchors and sacrifices progress without further safety gains.}
\label{fig:sweep_lambda}
\vspace{-10pt}
\end{figure}

\textbf{Trust-region weight.}
Fig.~\ref{fig:sweep_lambda} shows the effect of the trust-region weight $\lambda$, which penalizes deviations of the refined trajectory from its anchor. Without the trust region ($\lambda=0$), CEM can exploit the predicted scene more aggressively, leading to the highest progress on both benchmarks (e.g., $+4.0$ on Test14-hard relative to the default setting). However, this also results in overconfident refinements: time-to-collision compliance (TTC) decreases by $6.3$ on Test14-hard and $6.7$ on interPlan-LC. This behavior is expected, the agent predictions are conditioned on the anchor trajectory and are therefore only reliable within a local neighborhood. 

In contrast, a large trust-region weight ($\lambda=0.5$) constrains the refinement too strongly to the anchor. This reduces progress ($-2.5$ on Test14-hard) without yielding further safety improvements, as the optimizer has insufficient freedom to improve the original trajectory. This result supports our motivation: allowing moderate deviations from the anchor improves progress while safety
remains unaffected. The default setting of $\lambda{=}0.1$ provides the best trade-off and achieves the highest overall score on both benchmarks. These results demonstrate the role of the trust region as a regularizer for prediction validity: it allows local refinement of the ego trajectory while preventing large deviations for which the anchor-conditioned prediction may no longer be accurate.

\textbf{Component ablation.} 
\begin{table}[t]
\centering
\footnotesize
\setlength{\tabcolsep}{3pt}
\caption{Component ablation on Test14-hard and the lane-change scenarios of interPlan (interPlan-LC). Each row removes one component of the full approach and reports the difference to the full approach (blue: better, red: worse).}
\label{tab:component-ablation}
\resizebox{\columnwidth}{!}{%
\begin{tabular}{@{}lrrrrrrr@{}}
\toprule
Configuration & \shortstack{At-\\Fault} & TTC & DAC & \shortstack{Making\\Progress} & \shortstack{Progress\\along Route} & Comfort & Score \\
\midrule
\multicolumn{8}{c}{\textit{Test14-hard}} \\
\midrule
\textbf{Ours} & 97.43 & 89.71 & 98.90 & 96.32 & 85.10 & 81.25 & 85.35 \\
\quad w/o anchors & \dneg{$-$4.25} & \dneg{$-$10.00} & \dneg{$-$4.43} & \dneg{$-$5.92} & \dneg{$-$4.17} & \dpos{+12.48} & \dneg{$-$13.35} \\
\quad w/o refinement & \dpos{+0.74} & \dpos{+2.94} & \dneg{$-$0.37} & \dneg{$-$1.47} & \dneg{$-$7.40} & \dneg{$-$2.57} & \dneg{$-$2.92} \\
% \quad w/o agent-aware prior & \dneg{$-$5.33} & \dneg{$-$11.03} & \dzero{0.00} & \dpos{+0.37} & \dpos{+1.43} & \dpos{+6.99} & \dneg{$-$5.82} \\
\quad w/o continuous score & \dneg{$-$1.47} & \dneg{$-$3.31} & \dzero{0.00} & \dpos{+0.37} & \dneg{$-$1.16} & \dpos{+0.74} & \dneg{$-$2.19} \\
\midrule
\multicolumn{8}{c}{\textit{interPlan-LC}} \\
\midrule
\textbf{Ours} & 100.00 & 100.00 & 100.00 & 100.00 & 80.34 & 76.67 & 89.07 \\
\quad w/o anchors & \dneg{$-$3.33} & \dneg{$-$3.33} & \dneg{$-$13.33} & \dzero{0.00} & \dpos{+4.09} & \dpos{+13.33} & \dneg{$-$12.19} \\
\quad w/o refinement & \dzero{0.00} & \dneg{$-$3.33} & \dzero{0.00} & \dzero{0.00} & \dneg{$-$6.60} & \dneg{$-$13.33} & \dneg{$-$4.95} \\
% \quad w/o agent-aware prior & \dzero{0.00} & \dneg{$-$6.67} & \dzero{0.00} & \dzero{0.00} & \dpos{+2.54} & \dneg{$-$16.67} & \dneg{$-$4.41} \\
\quad w/o continuous score & \dzero{0.00} & \dneg{$-$6.67} & \dzero{0.00} & \dzero{0.00} & \dneg{$-$2.22} & \dpos{+10.00} & \dneg{$-$3.07} \\
\bottomrule
\end{tabular}}
\vspace{-5pt}
\end{table}

Tab.~\ref{tab:component-ablation} analyzes the contribution of the individual components of INTERACT on Test14-hard and interPlan-LC. Removing any component decreases the overall score on both benchmarks, indicating that each component addresses important aspects.

The anchor trajectories constitute the most important component. Removing them causes the largest performance drop, with $-13.35$ on Test14-hard and $-12.19$ on interPlan-LC. On Test14-hard, this degradation is reflected across both safety and progress metrics, including $-10.00$ TTC, $-4.25$ At-Fault, and $-5.92$ Making Progress. Similarly, on interPlan-LC, drivable-area compliance (DAC) decreases by $-13.33$. Interestingly, comfort improves without anchors on both benchmarks. This suggests that optimizing without anchors can produce locally smooth or progressive trajectories, but these trajectories are less reliable with respect to interaction safety and map constraints. The anchors therefore provide a structured prior over plausible driving intents.

Removing the CEM refinement results in a smaller but consistent decrease of $-2.92$ on Test14-hard and $-4.95$ on interPlan-LC. On Test14-hard, the unrefined anchors are slightly safer in terms of at-fault collision (At-Fault) and TTC, but lose $-7.40$ in progress. The effect is even clearer on interPlan-LC, where removing refinement reduces route progress by $-6.60$ and comfort by $-13.33$. This indicates that the anchor set already provides strong and conservative driving intents, while the refinement step is primarily responsible for adapting these trajectories to the predicted interaction and improving their efficiency and quality.

Finally, removing the continuous score reduces performance by $-2.19$ on Test14-hard and $-3.07$ on interPlan-LC. The most consistent degradation occurs in TTC ($-3.31$ and $-6.67$, respectively), accompanied by reduced route progress on both benchmarks. In contrast to discrete scoring, the continuous formulation provides a more informative ranking of CEM candidates and allows the optimizer to distinguish between trajectories with different degrees of safety and progress even when they satisfy the same discrete criterion. Overall, the ablation shows that the anchors provide the structural basis of INTERACT, while the continuous scoring, and local CEM refinement progressively improve interaction safety and driving efficiency.

\begin{table}[t]
\centering
\footnotesize
\setlength{\tabcolsep}{3pt}
\caption{Effect of prediction used on Test14-hard and lane-change scenarios of interPlan (interPlan-LC). Rows below ours report the difference to our anchor-conditioned SMART prediction (blue: better, red: worse).}
\label{tab:prediction-ablation}
\resizebox{\columnwidth}{!}{%
\begin{tabular}{@{}lrrrrrrr@{}}
\toprule
Prediction & \shortstack{At-\\Fault} & TTC & DAC & \shortstack{Making\\Progress} & \shortstack{Progress\\along Route} & Comfort & Score \\
\midrule
\multicolumn{8}{c}{\textit{Test14-hard}} \\
\midrule
\textbf{Conditioned} (Ours) & 97.43 & 89.71 & 98.90 & 96.32 & 85.10 & 81.25 & 85.35 \\
Unconditioned & \dpos{+0.37} & \dpos{+0.74} & \dzero{0.00} & \dneg{$-$0.37} & \dzero{0.00} & \dpos{+2.21} & \dpos{+0.14} \\
Constant velocity & \dneg{$-$0.37} & \dneg{$-$1.47} & \dzero{0.00} & \dzero{0.00} & \dneg{$-$0.19} & \dneg{$-$1.47} & \dneg{$-$1.24} \\
\midrule
\multicolumn{8}{c}{\textit{interPlan-LC}} \\
\midrule
\textbf{Conditioned} (Ours) & 100.00 & 100.00 & 100.00 & 100.00 & 80.34 & 76.67 & 89.07 \\
Unconditioned & \dzero{0.00} & \dzero{0.00} & \dzero{0.00} & \dzero{0.00} & \dneg{$-$6.12} & \dzero{0.00} & \dneg{$-$4.36} \\
Constant velocity & \dzero{0.00} & \dneg{$-$3.33} & \dzero{0.00} & \dzero{0.00} & \dneg{$-$2.05} & \dzero{0.00} & \dneg{$-$3.79} \\
\bottomrule
\end{tabular}}
\vspace{-15pt}
\end{table}

\textbf{Predictions.} Tab.~\ref{tab:prediction-ablation} studies effects of predictions during trajectory scoring. We compare our anchor-conditioned prediction against unconditioned learned prediction and a constant-velocity baseline. While all three provide similar performance on Test14-hard, larger differences emerge on the more interaction-heavy interPlan-LC benchmark.

On Test14-hard, the unconditioned predictor performs nearly identically to the conditioned variant and achieves a marginally higher overall score ($+0.14$). This suggests that, for many scenarios in this benchmark, modeling the future motion of surrounding agents is more important than explicitly conditioning their behavior on the ego trajectory. In contrast, constant-velocity predictions reduce the overall score by $-1.24$, together with lower TTC compliance ($-1.47$) and comfort ($-1.47$), indicating that learned predictions provide a more accurate representation of the surrounding traffic even when explicit interaction modeling is less critical.

The benefit of conditioning becomes more pronounced on interPlan-LC. Replacing the conditioned prediction with the unconditioned variant reduces the overall score by $-4.36$, primarily due to a $-6.12$ reduction in progress. Constant-velocity predictions similarly decrease the score by $-3.79$ and additionally reduce TTC compliance by $-3.33$. These results are consistent with the highly interactive nature of lane-change scenarios: the future behavior of surrounding agents depends on the intended ego maneuver, and a prediction that is independent of the ego trajectory cannot capture such reactions. Conditioning the prediction on the anchor therefore provides the planner with an interaction-consistent estimate of the future scene.

Overall, the results indicate that reactive conditioning is most beneficial in scenarios where the ego vehicle actively influences the behavior of other agents. In less interactive settings, unconditioned learned predictions can already provide a strong estimate of future traffic evolution, whereas fixed constant-velocity predictions consistently lose performance due to their inability to model more complex agent behavior.

\begin{figure}[t]
\centering
\includegraphics[width=0.9\columnwidth]{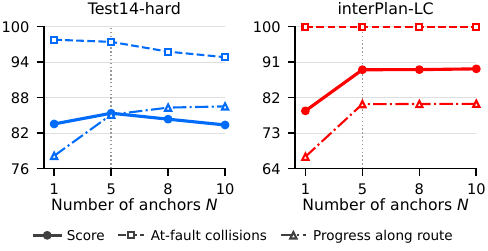}
\vspace{-6pt}
\caption{Sensitivity to the number of anchors $N$ on Test14-hard (left) and interPlan-LC (right): closed-loop score (solid), at-fault collisions (dashed) and progress along the route (dash-dotted). Going from one to five anchors unlocks progress additional anchors only add risk (at-fault collisions on Test14-hard) without improving the score.}
\label{fig:anchor_sweep}
\vspace{-10pt}
\end{figure}

\textbf{Anchors.} Fig.~\ref{fig:anchor_sweep} analyzes the sensitivity of INTERACT to the number of anchors $N$. Increasing the number of anchors from one to five substantially improves performance on both benchmarks. With only a single anchor, the planner lacks sufficient maneuver diversity, which mainly limits progress, particularly on interPlan-LC. At $N=5$, the anchor set captures the relevant maneuver alternatives, leading to a clear increase in progress and the highest overall scores. Adding further anchors provides little additional benefit. On interPlan-LC, both progress and score saturate, indicating that the required lane-change intents are already covered. On Test14-hard, additional anchors slightly increase progress but reduce the overall score due to a higher rate of at-fault collisions. This suggests that overly large anchor sets mainly introduce additional, potentially riskier maneuver alternatives rather than useful new intents. Overall, $N=5$ provides the best trade-off between maneuver diversity and robustness.

\subsection{Runtime Analysis}
\begin{table}[t]
\centering
\footnotesize
\setlength{\tabcolsep}{6pt}
\caption{Runtime (ms) per anchor averaged over ten interactive scenarios.}
\label{tab:runtime-breakdown}
\begin{tabular}{@{}lr@{}}
\toprule
Stage & Runtime (ms) \\
\midrule
Anchor generation & 32.6 \\
SMART inference   & 17.2 \\
Refinement        & 31.8 \\
\bottomrule
\end{tabular}
\vspace{-15pt}
\end{table}
Tab.~\ref{tab:runtime-breakdown} reports the runtime of a single anchor, as
all anchors are processed independently and can therefore be evaluated in
parallel. The reported SMART inference time corresponds to a single token,
i.e.\ one rollout step of the prediction model, which has to be queried
repeatedly to cover the full horizon. We use ten tokens ($5\,\mathrm{s}$) by
default: the planned trajectory spans $4\,\mathrm{s}$, and the additional
second is required to evaluate time-to-collision at the end of the horizon.
Shortening the horizon reduces the inference cost proportionally and leads to a slight drop in performance. 
Beyond that, the overall runtime is dominated by the prediction model and can be reduced further by replacing
SMART with a faster predictor or reducing the parameters.

\section{Conclusion and Future Work}

We propose INTERACT, an interactive planning framework that combines anchor-conditioned prediction with trust-region trajectory refinement. By separating prediction across maneuver intents from optimization within each intent, INTERACT incorporates reactive learned predictions without repeatedly querying the prediction model during optimization. The anchors provide diverse maneuvers, while local CEM refinement adapts them to the predictions. Experiments on the nuPlan closed-loop benchmark show strong performance across a wide range of scenarios, with large gains on challenging and highly interactive benchmarks.

Future work may extend INTERACT with adaptive trust regions that adjust the refinement range based on sensitivity of surrounding-agent predictions to changes in the ego trajectory. Another direction is to apply the refinement to end-to-end driving models, using their predicted trajectories as anchors and refining them with a reactive prediction model.

%%%%%%%%%%%%%%%%%%%%%%%%%%%%%%%%%%%%%%%%%%%%%%%%%%%%%%%%%%%%%%%%%%%%%%%%%%%%%%%%
%\clearpage

\addtolength{\textheight}{-12cm}   % This command serves to balance the column lengths

\bibliographystyle{IEEEtran}
\bibliography{main}

\end{document}